\documentclass{article}

\PassOptionsToPackage{numbers}{natbib}
\usepackage[final]{bdl_2021_camera_ready}

\usepackage[utf8]{inputenc} 
\usepackage[T1]{fontenc}    
\usepackage{hyperref}       
\usepackage{url}            
\usepackage{booktabs}       
\usepackage{amsfonts}       
\usepackage{nicefrac}       
\usepackage{microtype}      
\usepackage{xcolor}         
\usepackage{microtype}
\usepackage{graphicx}
\usepackage{subfigure}
\usepackage{amsmath, amssymb}
\usepackage{booktabs}
\usepackage{multirow}
\usepackage{comment}
\usepackage{caption}
\usepackage{varwidth}

\usepackage{float}
\DeclareMathOperator*{\argmax}{arg\,max}

\usepackage{hyperref}

\usepackage{bbm}
\usepackage{graphicx}

\usepackage[export]{adjustbox}
\graphicspath{ {./figures/} }
\usepackage{soul}

\title{Reproducible, incremental representation learning with Rosetta VAE}

\author{%
  Miles Martinez\\
  Electrical \& Computer Engineering \\ 
  Center for Cognitive Neuroscience\\
  Duke University\\
  \texttt{miles.martinez@duke.edu} \\
  \And 
  John Pearson \\
  Biostatistics \& Bioinformatics \\
  Center for Cognitive Neuroscience\\
  Electrical \& Computer Engineering \\
  Neurobiology \\
  Psychology \& Neuroscience \\
  Duke University \\
  \texttt{john.pearson@duke.edu} \\
}

\begin{document}

\maketitle

\section{Introduction}
Variational autoencoders are among the most popular methods for distilling low-dimensional structure from high-dimensional data, making them increasingly valuable as tools for data exploration and scientific discovery. However, unlike typical machine learning problems in which a single model is trained once on a single large dataset, scientific workflows privilege learned features that are reproducible, portable across labs, and capable of incrementally adding new data. Ideally, methods used by different research groups should produce comparable results, even without sharing fully-trained models or entire data sets. Here, we address this challenge by introducing the Rosetta VAE (R-VAE), a method of distilling previously learned representations and retraining new models to reproduce and build on prior results. The R-VAE uses post hoc clustering over the latent space of a fully-trained model to identify a small number of Rosetta Points (input, latent pairs) to serve as anchors for training future models. An adjustable hyperparameter, $\rho$, balances fidelity to the previously learned latent space against accommodation of new data. We demonstrate that the R-VAE reconstructs data as well as the VAE and $\beta$-VAE, outperforms both methods in recovery of a target latent space in a sequential training setting, and dramatically increases consistency of the learned representation across training runs. 

\section{Related Work}
Our approach is conceptually related to several strands of recent work. First, the notion of distilling a dataset by means of a small number of representative points is often studied under the heading of coresets \citep{agarwal2005geometric,feldman2011unified,sener2017active, borsos2020coresets} or Bayesian coresets \citep{huggins2016coresets,mak2018support,campbell2019automated,campbell2019sparse}. 
Our method is simpler in that we use standard clustering to determine our data subset, an approach that leverages a large body of research on scalable clustering. 

Second, this work intersects with recent results in identifiability for VAEs \citep{khemakhem2020,sorrenson2020disentanglement,zhou2020learning,kumar2020bvaeReg}. 
Our approach, while providing fewer guarantees than these results, appears to work without these assumptions, since it replaces constraints on the encoding model class with a set of point constraints that approximately identify the latent space. 

Third, the R-VAE shares with the VQ-VAE \citep{vandenoordDiscrete2017, razavi2019generating} in both its hard and soft versions \citep{sonderby2017continuous, roy2018theory, henter2018deep,wu2020vector, dib2020quantized} the notion of a quantization of latent space. The major distinction between these approaches and ours is that the VQ-VAE and its variants employ quantization as a regularization strategy for avoiding posterior collapse, while here our focus is on issues of portability and reproducibility. 

Finally, the portability problem as we describe it is closely related to those addressed by both continual learning and transfer learning. While in continual learning, the focus is often on learning new tasks, several studies have used distillation, including coresets, as an intermediate step in this process (e.g., \cite{shin2017, nguyen2018,schwarz2018, borsos2020coresets}). 

\begin{figure}[H]
    \centering
    \includegraphics[width=0.8\linewidth]{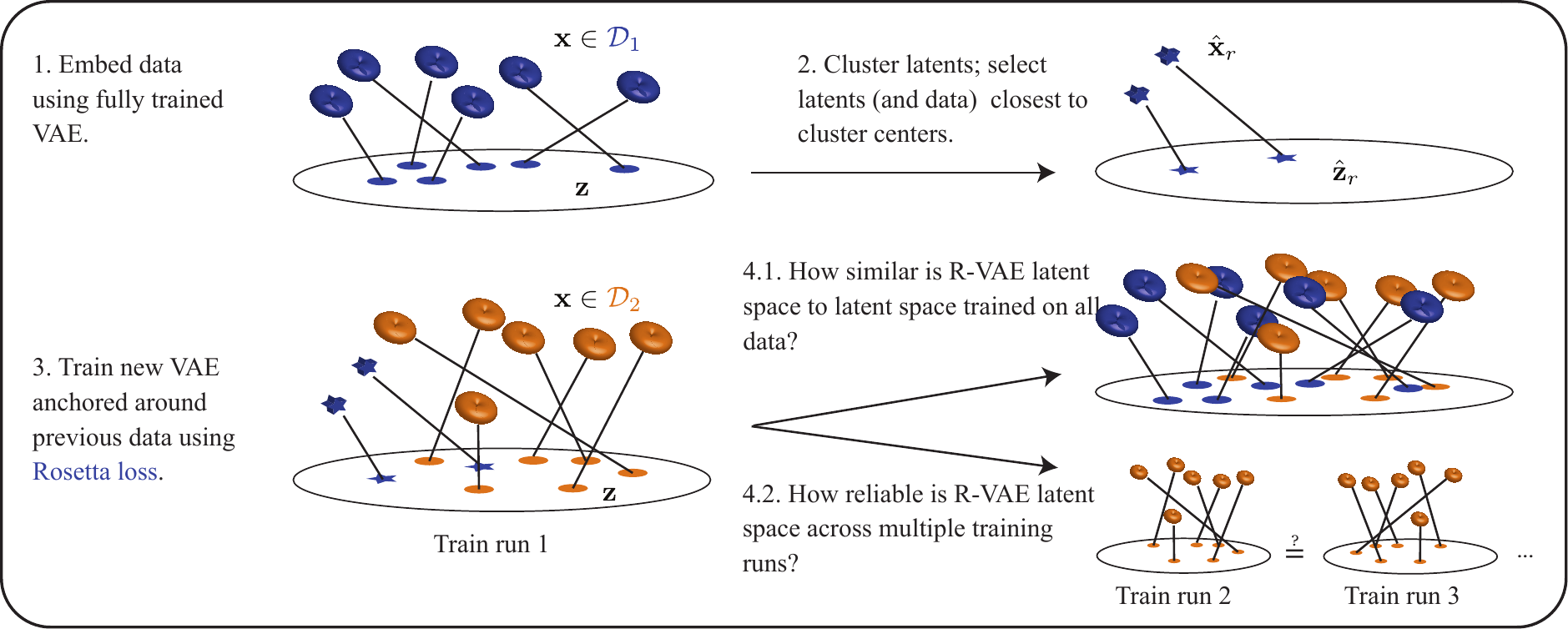}
    \caption{Schematic of Rosetta VAE training. \textbf{1)} Beginning with a trained VAE, data $\mathbf{x} \in \mathcal{D}_1$ are encoded to their latent representations $\mathbf{z}$. \textbf{2)} After clustering the latent representations, we identify Rosetta points $\mathbf{\hat{z}}_r$ closest to the cluster centroids and their preimages in the training set, $\mathbf{\hat{x}}_r$. \textbf{3)} For new data $\mathcal{D}_2$, the standard ELBO  is augmented with a loss term (\ref{eqn:rosetta_objective}) to enforce the constraint that the VAE preserves the Rosetta points $(\mathbf{\hat{x}}_r, \mathbf{\hat{z}}_r)$. \textbf{4.1)} For sequential training, we assess the similarity between the latent space inferred using the R-VAE and the latent space found by training jointly on all data. \textbf{4.2)} For reproducible training, we use Rosetta points from $\mathcal{D}_1$ to seed retraining on $\mathcal{D}_1$ and assess the reliability of the resulting embeddings across repeated training runs.} 
    \label{fig:f1}
\end{figure}
\section{Experiments}
\subsection*{Experiment structure and metrics}
We focus on the two experimental settings, the sequential and reproducible training problems. In the former, we consider two scientists, Alice and Bob, who wish to embed their data in a shared latent space. In the portability setting, we ask how similar the latent space discovered by Bob --- who has no access to Alice's data, or even Alice's trained model --- can be made to the one found by joint training on both data sets combined. In the reproducibility case, we ask how similar latent embeddings of Alice's data can be made across repeated retrainings, as the variability of latent spaces learned over retraining has been previously demonstrated \citep{locatello2019,sikka2019closer,khemakhem2020}. Our goal is therefore to enforce similarity between a previously learned latent space and a new latent space. We do so through the process outlined in Figure~\ref{fig:f1}: First, we embed data using a fully trained VAE (trained on $\mathcal{D}_1$), then distill that latent space through standard clustering into a small set of \textbf{Rosetta Points} (latent, data pairs). We then train a new VAE anchored around our Rosetta points using our \textbf{Rosetta loss}:

\begin{equation}
    \label{eqn:rosetta_objective}
    \mathcal{L}_\rho = \mathcal{L}(\theta, \phi) - \rho \sum_{r=1}^R \Bigg[\lVert \mathbf{\hat{x}}_r - \mathbf{m}(\mathbf{\hat{z}}_r)\rVert^2 + \lVert\mathbf{\hat{z}}_r - \boldsymbol{\mu}(\mathbf{\hat{x}}_r)\rVert^2 \Bigg] 
\end{equation}
where the $\mathbf{\hat{z}}_r$ are chosen to be the closest points to the centroids found by clustering $\mathbb{E}_{\mathbf{x}\sim \mathcal{D}_1}q_{\phi_*}(\mathbf{z}|\mathbf{x})$ and $\mathcal{L}(\theta,\phi)$ is the standard ELBO. In the reproducibility case, $\mathcal{L}_\rho(\theta,\phi)$ is optimized over $\mathcal{D}_1$ again. In the sequential training case, $\mathcal{L}_\rho(\theta,\phi)$ is optimized over a separate dataset, $\mathcal{D}_2$, and Rosetta points from $\mathcal{D}_1$.

To assess the similarity of latent spaces across training, we introduce two new metrics. For the sequential training setting, as in \cite{khemakhem2020}, we calculate a normalized distortion that considers the two latent spaces the same if they differ only by a linear map. That is, for a given data set $\mathcal{D}$ and a pair of encoders $q$ and $q'$, we let $\boldsymbol{\mu}(\mathbf{x}) = \mathbb{E}_{q(\mathbf{z}|\mathbf{x})} \mathbf{z}$ (with a similar definition for $\boldsymbol{\mu'}$) and calculate the \textbf{latent space distortion} as
\begin{equation}
    \label{eqn:distortion}
    LSD = \min_{\mathbf{A}, \mathbf{b}} \mathbb{E}_{\mathbf{x} \in \mathcal{D}} \lVert \mathbf{A}\cdot \boldsymbol{\mu}(\mathbf{x}) + \mathbf{b} - \boldsymbol{\mu'}(\mathbf{x}) \rVert^2 ,
\end{equation}
where $\mathbf{A}$ and $\mathbf{b}$ parameterize a linear map. For our sequential training experiments, $q'$ is the embedding learned by joint training on the full data set $\mathcal{D} = \mathcal{D}_1 \cup \mathcal{D}_2$ and $q$ is the encoder learned by training the R-VAE and comparison models on $\mathcal{D} = \mathcal{R}_1 \cup \mathcal{D}_2$. Note that this measure of distortion uses only the mean of the embedding map, ignoring uncertainty in the mapping from $\mathbf{x}$ to $\mathbf{z}$. We choose not to use mean correlation coefficient as in \citep{khemakhem2020}, since we do not know the "true" latent space.

For reproducible training, we measure \textbf{retraining variability}, which we defined as the average volume of the covariance matrix across training runs for each data point. If we label encoders learned by sequential training runs $1 \ldots M$ as $q_1 \ldots q_M$ with conditional means $\boldsymbol{\mu}_1 \ldots \boldsymbol{\mu}_M$, then
\begin{equation}
    \label{eqn:retraining_variability}
    RV = \mathbb{E}_{\mathbf{x} \in \mathcal{D}} \log \det \mathbf{C}(\mathbf{x}), \qquad \mathbf{C(x)} = \mathrm{cov}\left(
    \begin{bmatrix}
        \boldsymbol{\mu}_1(\mathbf{x}) & \boldsymbol{\mu}_2(\mathbf{x}) & \ldots & \boldsymbol{\mu}_M(\mathbf{x}) 
    \end{bmatrix}
    \right) . 
\end{equation}
This index then gives a (logged) volume measure of the ellipsoid containing the same data point's latent representation across runs, with lower numbers indicating more reliable embeddings. See appendix for datasets used and model training details.


\subsection{Rosetta VAE consistently recovers the same latent space across retrainings}
\label{sec:repro}
To test the ability of the Rosetta VAE training procedure to reproduce consistent latent representations of the same data across retrainings, we first trained a standard VAE to embed each of our example datasets. We then clustered the resulting (mean) latent representations using $k$-means with $k=8$ for the Gaussians and $k=64$ for MNIST and Birdsong, taking the closest data embedding to each cluster centroid ($\mathbf{\hat{z}}_r$) and its associated data point ($\mathbf{\hat{x}}_r$) as the Rosetta points ($\mathcal{R}_1$).  We then used $\mathcal{R}_1$ and the original dataset to retrain a VAE, $\beta$-VAE, and R-VAE 10 times each and assessed the consistency of the embeddings across runs using (\ref{eqn:retraining_variability}). Note that, except for the R-VAE, only the $\mathbf{\hat{x}}_r$ from $\mathcal{R}_1$ were used. That is, we duplicated a small number of points from the original data set without upweighting them or fixing their embedding locations.

As visualized in Figure \ref{fig:f2} and quantified in Table \ref{table:t1}, the R-VAE produced much more consistent embeddings across training runs. In fact, this was true even when retraining the standard VAE using the same initial seed, due to GPU nondeterminism. 
Thus, retraining with the Rosetta VAE allowed us to reproduce the same latent space structure again and again by seeding only with a handful of latent data points and their embeddings. 

\begin{table}[ht]
\begin{minipage}{0.45\textwidth}
\centering
\captionof{figure}{}

\includegraphics[width=\linewidth]{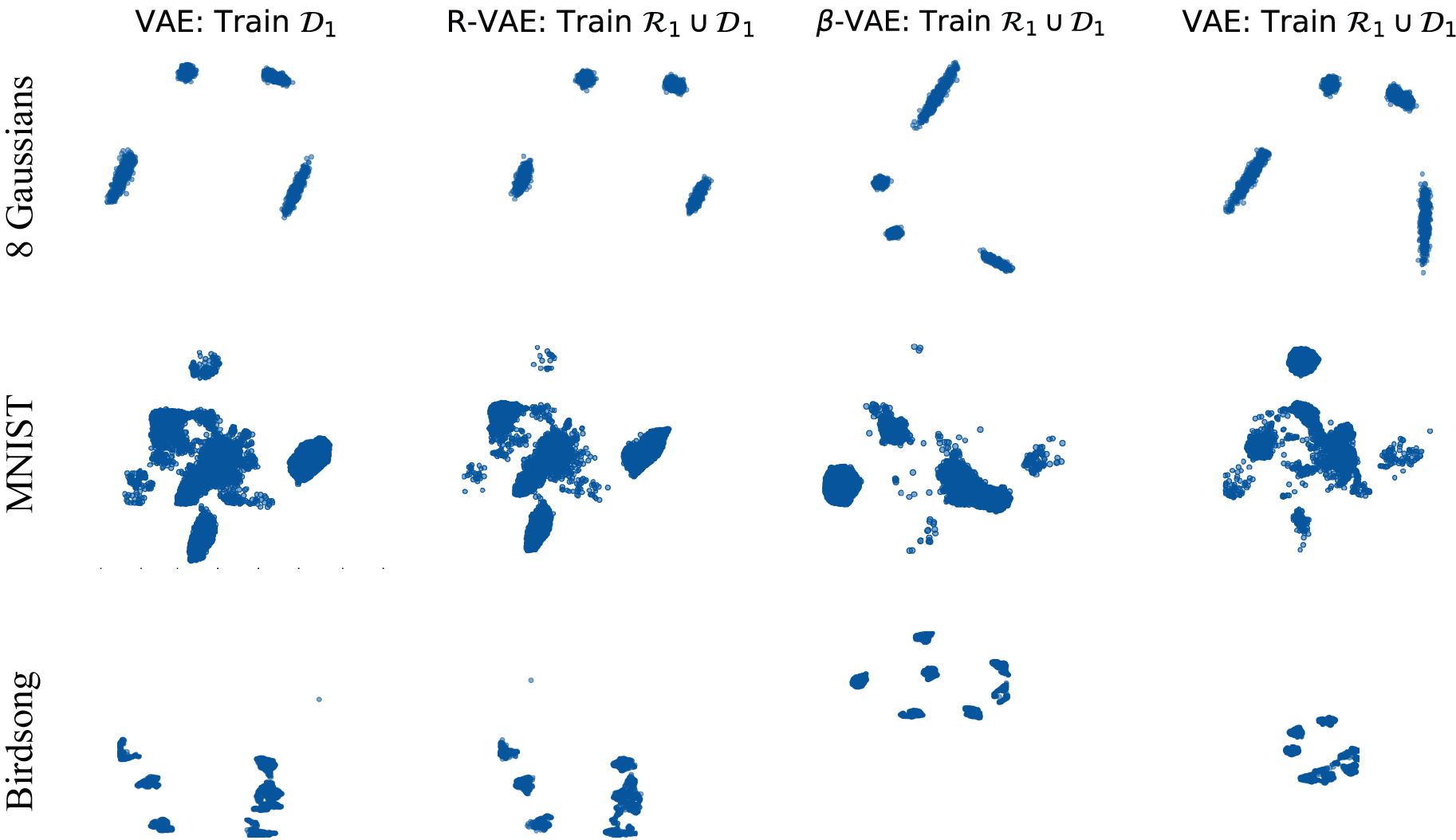}

\label{fig:f2}

\end{minipage}
\hfill
\begin{minipage}{0.5\textwidth}
\centering
\renewcommand{\arraystretch}{1.5}
\caption{}

\resizebox{\linewidth}{!}{%
    \begin{tabular}{lllll}
    \toprule
    & & 8 Gaussians & MNIST & Birdsong \\ \toprule
    \multirow{5}{*}{RV($\mathcal{R}_1$)} &&&& \\
    & VAE& 0.00(0.193) & 0.00 (5.861) & 0.00(14.512) \\ 
    &$\beta$-VAE& -1.151(0.970) & -4.934478(8.023) & -33.568 (27.799)\\ 
    &VAE (same seed)& --- & -48.828(9.352) & -9.123(11.649) \\ 
    & R-VAE & \textbf{-17.247(1.496)} &  \textbf{-95.393 (5.268)} & \textbf{-149.942(12.198)}\\ \cmidrule{1-5}
    \multirow{5}{*}{RV(rest of $\mathcal{D}_1$)} & &&& \\ 
    &VAE& 0.00(0.355) & 0.00(14.254) & 0.00(17.015) \\ 
    &$\beta$-VAE& -1.204(1.003) & -5.671(10.069) & -40.019(22.808)\\ 
    &VAE (same seed)& --- & \textbf{-48.566(11.05)} & -7.542(12.147)\\ 
    & R-VAE & \textbf{-16.425(1.840)} & \textbf{-46.241(12.593)} &  \textbf{-82.440(23.285)} \\ \bottomrule
    \end{tabular}}
\label{table:t1}

\end{minipage}
\end{table}
\textbf{Figure 2, Table 1: R-VAE reproduces the same latent space across training runs.} (left panel) Learned latent representations of each dataset, projected to two dimensions by UMAP, in \textbf{(first column)} and as reproduced by retraining with Rosetta VAE, standard VAE, and  $\beta$-VAE \textbf{(columns 2--4)}. $R=8,64,64$  Rosetta points used for 8 Gaussians, MNIST, and birdsong respectively. In each case, R-VAE matches the target latent space, consistently across retrainings. (right panel) Medians and interquartile ranges of retraining variability metric for three example data sets, normalized by median vanilla VAE performance.
\captionsetup[table]{labelsep=colon}

\subsection{Rosetta VAE approximates full data latent spaces under sequential training}\label{sec:seq}
To assess the ability of the Rosetta VAE procedure to capture the full data latent space under sequential training, we partitioned each of our test data sets into halves as specified in Appendix \ref{sec:datasets}: $\mathcal{D} = \mathcal{D}_1 \cup \mathcal{D}_2$. We then performed VAE training as described above on $\mathcal{D}_1$, distilled the mean latent embeddings of these data via $k$-means clustering ($k=8$ for the 8 Gaussians, $k=32$ for Birdsong, $k=64$ for MNIST, and $k=128$ for 3D Chairs) to produce a set of Rosetta points $\mathcal{R}_1$. These points were then used to train an R-VAE model on $\mathcal{R}_1 \cup \mathcal{D}_2$ using the Rosetta Loss (\ref{eqn:rosetta_objective}). We then assessed the similarity of the latent spaces found by this procedure to the latent space trained on $\mathcal{D}_1 \cup \mathcal{D}_2$ via the distortion measure (\ref{eqn:distortion}), which we report separately for both $\mathcal{D}_1$ and $\mathcal{D}_2$ in Table \ref{table:t2} (using the same linear map for both). See Appendices \ref{app:sequential_latents_correct} and \ref{app:sequential_latents}  for visualization of embeddings in each case.

\begin{table}[h]
    \caption{Medians and interquartile ranges of distortion for four example data sets, normalized by median vanilla VAE performance.}
    \centering
    \resizebox{\textwidth}{!}{%
    \begin{tabular}{llllllll}\toprule
    & \multicolumn{6}{c}{Latent Space Distortion} \\\cmidrule(lr){2-7}
    & \multicolumn{3}{c}{$\mathcal{D}_1$} &  \multicolumn{3}{c}{$\mathcal{D}_2$}
    \\\cmidrule(lr){2-4}\cmidrule(lr){5-7} 
         & VAE & $\beta$-VAE & R-VAE & VAE & $\beta$-VAE & R-VAE  \\\midrule 
    8 Gaussians & 0.00(0.216)  & \textbf{-0.056(0.066)}  &  \textbf{-0.049(0.057)} & 0.00(0.304) & -0.060(0.336) & \textbf{-0.303(0.056)} \\
    MNIST & 0.00 (0.049)& 0.989(0.656)& \textbf{-0.749(0.220)}& 0.00(0.431) & 0.913(0.754)& \textbf{-0.982(0.069)} \\
    Birdsong  & 0.00 (0.541)& -0.379(0.326)& \textbf{-0.512(0.116)}  & 0.00 (0.376) & \textbf{-0.588(0.305)} & -0.153(0.529)  \\
    3D Chairs & \textbf{0.00(0.190)}& 3.509(0.436) & \textbf{-0.039(0.265)} & \textbf{0.00(0.213)}& 3.565(0.440) & \textbf{-0.033 (0.289)} \\
\bottomrule
\end{tabular}}
\label{table:t2}

\end{table}

We further examined the latent embeddings by asking how distorted latent maps were across the models and training runs, assessed by the similarity of the learned \textbf{A} matrix in (\ref{eqn:distortion}) to the identity (Appendix \ref{app:mapAnalysis}). Overall, the R-VAE required linear maps that were closer to the identity than the VAE and $\beta-$VAE, indicating less non-uniform stretching and compression of the learned space relative to the joint training template. The biases \textbf{b} are likewise small (Appendix \ref{app:learned_biases}).


\section{Discussion}
The issues of reproducibility and portability, while central to the scientific enterprise, have been sparsely addressed in the neural networks literature. Our Rosetta VAE provides a simple, intuitive prescription for increasing both. By adjusting a single parameter, we can trade off fidelity to previously learned latent spaces against accommodation of distribution shifts driven by new data. Perhaps surprisingly, latent representations for \emph{all} VAE models we tested showed striking reproducibility under sequential training, much more so than might be expected from representation learning results in, e.g., \cite{locatello2019}. The important qualification to this result, of course, is that when we assessed similarity between latent representations in the sequential training case, we calculated distortion modulo an overall linear transformation, similar to \cite{khemakhem2020}. This finding suggests that what might seem like large differences in learned representations in previous studies may simply be the result of linearly transformed latent variables. Of course, in the absence of a joint training template (Figure \ref{fig:sf0}, first column), this is impossible to assess.

Our Rosetta loss, which simply fixes the embedding locations of a small number of data points, is both simple to implement and conceptually intuitive: by ``tacking down'' our Rosetta points, we effectively remove symmetries in the latent space that prevent identifiability. In this sense, our $\mathcal{R}_1$ points are reminiscent of the exogenous covariates $\mathbf{u}$ that make identifiability possible in \cite{khemakhem2020}. Moreover, our results do not depend sensitively on either the number of Rosetta points nor the architecture of the model from which they are derived (Appendices \ref{app:RPnumber}, \ref{app:RPselection}, \ref{app:model_agnostic}), suggesting a robust, practical method for VAE reproducibility. However, as noted above, a limitation of this work is that we do not solve the coreset problem except heuristically and so offer none of the convergence guarantees of the coreset or identifiability literature (e.g., \cite{feldman2011unified,sener2017active, huggins2016coresets,khemakhem2020}). 

It is also important to note that, as a data distillation method, our work potentially compounds problems of bias and underrepresentation in existing datasets. That is, in selecting Rosetta points near areas of high density in latent space, it is most likely to preserve the most typical points in initial training sets most faithfully. As such, care must be taken not to exacerbate bias in sensitive applications, and more studies will be needed to assess its potential for harm. Conversely, our results on sequential training suggest that previously trained models may be augmented by new data without appreciable distortion of the latent space, which may help to redress problems in some models that result from gaps in training data.

Finally, our results confirm the practical utility of models like the $\beta$-VAE \cite{higgins2016beta} and regularization in general, in producing robust learning of latent spaces. While our R-VAE strongly outperformed both standard and $\beta$-VAEs in reproducing the same latent space across training runs, performance was more variable on sequential training. In cases like the 3D Chairs dataset, which we partitioned randomly, most models performed well, while the $\beta$-VAE outperformed R-VAE on $\mathcal{D}_2$ for the birdsong data, suggesting that dataset structure and the distribution shifts associated with adding new data to existing models may play a role. Further work is needed to combine the symmetry-breaking benefits of methods like the R-VAE with the overall regularization offered by $\beta$-VAE and related models.

\begin{ack}
We thank the Mooney Lab for birdsong data and Jack Goffinet for helpful conversations on data preprocessing and visualization. 
This work was funded by BRAIN grant R01-NS118424. 
\end{ack}

\bibliography{example_paper.bbl}
\bibliographystyle{plainnat}

\newpage
\appendix

\section{Datasets}
\label{sec:datasets}
For our experiments, we used four data sets of varying complexity: 1) a toy data set comprising 8 Gaussians in two dimensions, 2) MNIST (from \citet{lecun2010mnist}, licensed under Creative Commons Attribution-Share Alike 3.0 license), 3) 3D Chairs (from \citet{Aubry2014chairs}), and 4) spectrograms representing syllables of zebra finch birdsong (from \citet{Goffinet2021}, licensed under Creative Commons CC0 1.0 Universal). The last of these allows us to consider real data of a type that are known to exhibit strong clustering in latent space \cite{sainburg2020finding,Goffinet2021}. Moreover, they are of scientific interest because studies of birdsong both within and across individuals require multiple animals, and so latent space modeling should produce structures that are both reproducible and portable across labs. In each case, we partitioned the total data into sets $\mathcal{D}_1$ and $\mathcal{D}_2$ as follows: 1) four Gaussians each, separated by a half-plane; 2) Digits 0--4 and 5--9; 3) a random division of the data set into halves; 4) distinct sets of birds, with each individual singing only minor variants of a single song.

\section{Model Training}
\label{sec:modelTrain}
After division into $\mathcal{D}_1$ and $\mathcal{D}_2$ data sets as described above, data were further divided within each partition into 60/40 train/validation split. Hyperparameters $\beta$ and $\rho$ were selected by training models for 20 epochs at $\beta$ = [0:2.5:25], $\rho$ = [0:0.75:15]. The $\beta$ and $\rho$ with the best validation performance after this initial training were selected and used for experiments. For R-VAE training, $\rho$ was weighted by the ratio of the number of Rosetta Points to batch size, and \ref{eqn:rosetta_objective} was applied to the Rosetta Points alongside every batch of new data. Models were optimized using Adam \citep{kingma2015Adam} with learning rate 1e-3. Stopping criterion was determined by letting joint data models train until loss plateaued and then using that same number of epochs for each of the second-phase comparison models. For 8 Gaussians and MNIST we used 200 epochs, for all others we used 300 epochs. Appendix \ref{app:modelArchs} contains details of model architecture. All models were created and trained using PyTorch \citep{paszke2019pytorch} (licensed under BSD), code for all experiments in paper can be found in the supplemental material. Experiments on average required 20 hours to run on an RTX3070 GPU.

\section{Model Architectures}
\label{app:modelArchs}
\begin{figure}[H]
    \centering
    \includegraphics[width=\linewidth]{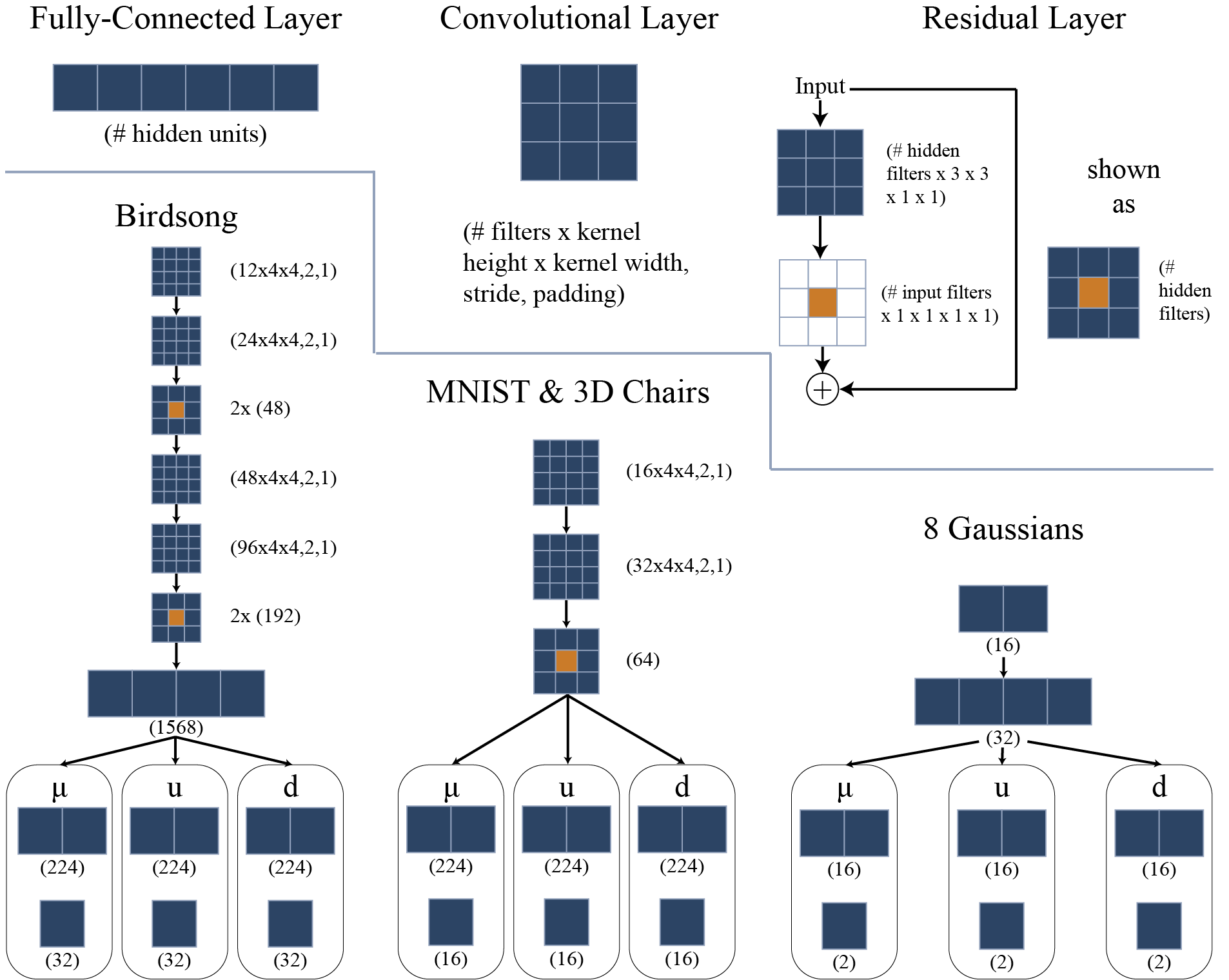}
    \caption{\textbf{Architectures used in main text.} \textbf{Top row}: types of layers used in the neural networks in the paper. Fully connected, convolutional, and residual layers were used. For fully connected layers, parameters used are presented underneath in parentheses; for convolutional and residual layers, parameters are shown beside the layer in parentheses. All residual layers had the same size hidden filters, and only differed in the number of hidden filters. \textbf{Bottom Row}: Encoder architectures used for each dataset in the paper, ordered by decreasing model complexity from left to right. MNIST \& 3d Chairs used the same model architectures, and differed only in input size. The input for birdsong was 128x128 spectrograms, for 3D Chairs 64x64 grayscale images, for MNIST 28x28 grayscale images, and for the 8 Gaussians 5-dimensional vectors (two (x,y) position dimensions + 5 dimensional gaussian noise). Output network $\mu$ parameterizes the mean $\mathbb{E}\mathbf{z}$, and networks $u$ and $d$ parameterize the (flattened) lower triangle and diagonal of the Cholesky decomposition of $\mathrm{cov}[\mathbf{z}]$, respectively. Only encoders are presented for space, all decoders were reverse of encoders, with convolutional layers replaced with transposed convolutional layers.}
    \label{fig:arch_fig}
\end{figure}

\section{Rosetta-VAE dependence on number of Rosetta Points}

Rosetta-VAE consistency did not depend strongly on the number of Rosetta Points. Experiments here followed the same form as those in the main text, with \textbf{reproducibility trained models} referring to the procedure in section \ref{sec:repro} and \textbf{sequentially trained models} to the procedure in section \ref{sec:seq}. Tables \ref{table:st1},\ref{table:st2},\ref{table:st3} show an extremely limited effect of the number of RPs, with even 4 RPs greatly improving the consistency of learned embeddings over VAE and $\beta$-VAE. The sequentially trained case shows slightly more reliance on number of RPs, but still remains consistent across a range of numbers of RPs.
\label{app:RPnumber}

\begin{table}[H]
\caption{Medians and interquartile ranges of RV for reproducibility trained 8 Gaussians, normalized by median vanilla VAE performance.}
\centering
\begin{tabular}{lllll}\toprule
& \multicolumn{4}{c}{RV} \\\cmidrule(lr){2-5} 
     & 2 RPs & 4 RPs & 8 RPs & 16 RPs  \\\midrule 
VAE & 0.00 (0.933)  & 0.00(0.316)  &  0.00(0.355) & 0.00(0.511) \\
$\beta$-VAE & \textbf{-1.225 (1.068)} & -1.021 (0.504)& -1.204 (1.00) & -0.934 (0.348)  \\
R-VAE  & \textbf{-1.412(4.191)} & \textbf{-14.656 (6.642)} & \textbf{-16.425 (1.840)}  & \textbf{-17.116 (1.649)}   \\

\bottomrule
\end{tabular}
\label{table:st1}
\end{table}

\begin{table}[H]
\caption{Medians and interquartile ranges of RV for reproducibility trained MNIST, normalized by median vanilla VAE performance.}
\centering

\begin{tabular}{lll}\toprule
& \multicolumn{2}{c}{RV} \\\cmidrule(lr){2-3} 
     & 32 RPs & 64 RPs \\\midrule 
VAE & 0.00 (9.557)  & 0.00(9.111) \\
$\beta$-VAE & -5.776 (10.243) & -5.671 (10.069) \\
R-VAE  & \textbf{-32.083 (15.603)} & \textbf{-46.241 (12.593)} \\

\bottomrule
\end{tabular}
\label{table:st2}
\end{table}

\begin{table}[H]
\caption{Medians and interquartile ranges of RV for reproducibility trained birdsong, normalized by median vanilla VAE performance.}
\centering
\resizebox{\textwidth}{!}{%
\begin{tabular}{llllll}\toprule
& \multicolumn{5}{c}{RV} \\\cmidrule(lr){2-6} 
     & 4 RPs & 8 RPs & 16 RPs & 32 RPs & 64 RPs  \\\midrule 
VAE & 0.00(14.254)  & 0.00(17.015)  &  0.00(15.805) & 0.00(18.083) & 0.00(16.672) \\
$\beta$-VAE & -46.503 (26.498)& -40.019 (22.808)& -38.892 (24.572) & -31.955 (27.672) & -41.852 (23.307) \\
R-VAE  & \textbf{-75.852 (20.358)} & \textbf{-82.440 (23.285)} & \textbf{-74.166 (27.809)}  & \textbf{-85.790 (32.503)} & \textbf{-95.529 (28.815)}   \\

\bottomrule
\end{tabular}}
\label{table:st3}
\end{table}

\begin{table}[H]
\caption{Medians and interquartile ranges of LSD for sequentially trained 8 Gaussians, normalized by median vanilla VAE performance.}
\centering
\begin{tabular}{llllll}\toprule
& \multicolumn{5}{c}{LSD} \\\cmidrule(lr){2-6} 
     & 2 RPs & 4 RPs & 8 RPs & 16 RPs & $\mathcal{D}_1$ \\\midrule 
VAE & \textbf{0.00 (0.115)}  & \textbf{0.00(0.024)}  &  0.00(0.216) & 0.00(0.283) & 0.00 (0.301) \\
$\beta$-VAE & 0.143 (0.293) & \textbf{0.00 (0.052)}& \textbf{-0.056 (0.066)} & \textbf{-0.042 (0.044)} & \textbf{-0.046 (0.042)}  \\
R-VAE  & 0.435 (0.355) & \textbf{0.015 (0.056)} & \textbf{-0.049 (0.057)} & \textbf{-0.029 (0.044)} & -0.007 (0.025)   \\

\bottomrule
\end{tabular}
\label{table:st4}
\end{table}

\begin{table}[H]
\caption{Medians and interquartile ranges of LSD for sequentially trained MNIST, normalized by median vanilla VAE performance.}
\centering

\begin{tabular}{llll}\toprule
& \multicolumn{3}{c}{LSD} \\\cmidrule(lr){2-4} 
     & 32 RPs & 64 RPs & $\mathcal{D}_1$ \\\midrule 
VAE & 0.00 (0.635)  & 0.00(0.487) & \textbf{0.00 (0.553)} \\
$\beta$-VAE & 0.904 (0.289) & 0.989 (0.656) & 1.217 (0.386) \\
R-VAE  & \textbf{-0.725 (0.113)} & \textbf{-0.749 (0.220)} & 0.770 (0.110) \\

\bottomrule
\end{tabular}
\label{table:st5}
\end{table}

\begin{table}[H]
\caption{Medians and interquartile ranges of LSD for sequentially trained birdsong, normalized by median vanilla VAE performance.}
\centering
\begin{tabular}{lll}\toprule
& \multicolumn{2}{c}{LSD} \\\cmidrule(lr){2-3} 
     & 32 RPs & 64 RPs \\\midrule 
VAE & 0.00(0.541)  & 0.00(0.394) \\
$\beta$-VAE & -0.379 (0.326)& \textbf{-0.588 (0.305)} \\
R-VAE  & \textbf{-0.512 (0.116)} & -0.153 (0.529) \\

\bottomrule
\end{tabular}
\label{table:st6}
\end{table}

\section{Rosetta-VAE is agnostic to Rosetta Point selection method}
Here, we compare R-VAE performance when different clustering methods are used to select the Rosetta points. We compare K-means, as used in the main text, to agglomerative clustering, Gaussian mixture model clustering, and random selection of embeddings. Data reported are from sequentially trained R-VAEs.
\begin{table}[H]
\caption{Medians and interquartile ranges of LSD \& RV for different RP selection methods in the sequential training setting, normalized by median R-VAE performance using K-Means clustering.}
\centering

\begin{tabular}{llll}
\toprule
& &  $\mathcal{D}_1$ & $\mathcal{D}_2$  \\ \toprule
\multirow{4}{*}{LSD} & && \\
& Agglomerative Clustering& \textbf{0.834 (2.209)} & \textbf{0.251 (2.115)}  \\ 
& Gaussian Mixture Model & \textbf{0.933 (2.204)} & \textbf{0.255 (2.130)}  \\ 
& K-Means & \textbf{0.00 (1.616)} & \textbf{0.00 (2.025)}   \\ 
& Random Selection &  \textbf{0.073 (1.698)} & 12.063 (8.485)    \\ \cmidrule{1-4}
\multirow{4}{*}{RV} & & & \\ 
& Agglomerative Clustering & 43.024 (34.094) & 17.636 (29.159)    \\ 
& Gaussian Mixture Model & 29.215 (24.532) & \textbf{-6.842 (20.999)}  \\ 
& K-Means & \textbf{0.00 (18.415)} & \textbf{0.00 (31.612)}   \\ 
& Random Selection & \textbf{9.356 (17.569)} & \textbf{4.367 (19.808)}    \\
 \bottomrule
\end{tabular}
\label{table:st7}
\end{table}

\label{app:RPselection}
\section{Rosetta-VAE is model agnostic}
R-VAEs were trained on birdsong using three different architectures. The first was used as a reference and had the same architecture as models used in the main text (and as the template model). Our ``complex'' model had an additional residual layer with 96 hidden units inserted between the third and fourth convolutional layers (see figure \ref{fig:arch_fig})  while our "simple" model had the first two existing residual layers removed. All models demonstrated similar performance in both metrics. Data reported are from sequentially trained R-VAEs.
\label{app:model_agnostic}

\begin{table}[H]
\caption{Medians and interquartile ranges of LSD \& RV for sequentially trained birdsong, normalized by performance of the R-VAE with the architecture of the template model.}
\centering
\resizebox{\textwidth}{!}{%
\begin{tabular}{lllllll}\toprule
& \multicolumn{3}{c}{$\mathcal{D}_1$} & \multicolumn{3}{c}{$\mathcal{D}_2$} \\\cmidrule(lr){2-4}  \cmidrule(lr){5-7}
     & Simple & Complex & Same Arch & Simple & Complex & Same Arch \\\midrule 
LSD & 0.175 (0.234) & 0.191 (0.273) & \textbf{0.00 (0.312)} & \textbf{-0.102 (0.147)} & 0.119 (0.171) & \textbf{0.00 (0.293)}
 \\\midrule 
RV & \textbf{1.485 (19.221)} & \textbf{3.912 (19.155)} & \textbf{0.00 (18.415)} & \textbf{-4.725 (27.337)} & \textbf{2.927 (30.421)} & \textbf{0.00 (31.612)} \\
\bottomrule
\end{tabular}}
\label{table:st8}
\end{table}
\section{Sequentially trained latent spaces corrected for linear transformation}
\label{app:sequential_latents_correct}
\begin{figure}[H]
    \centering
    \includegraphics[width=\linewidth]{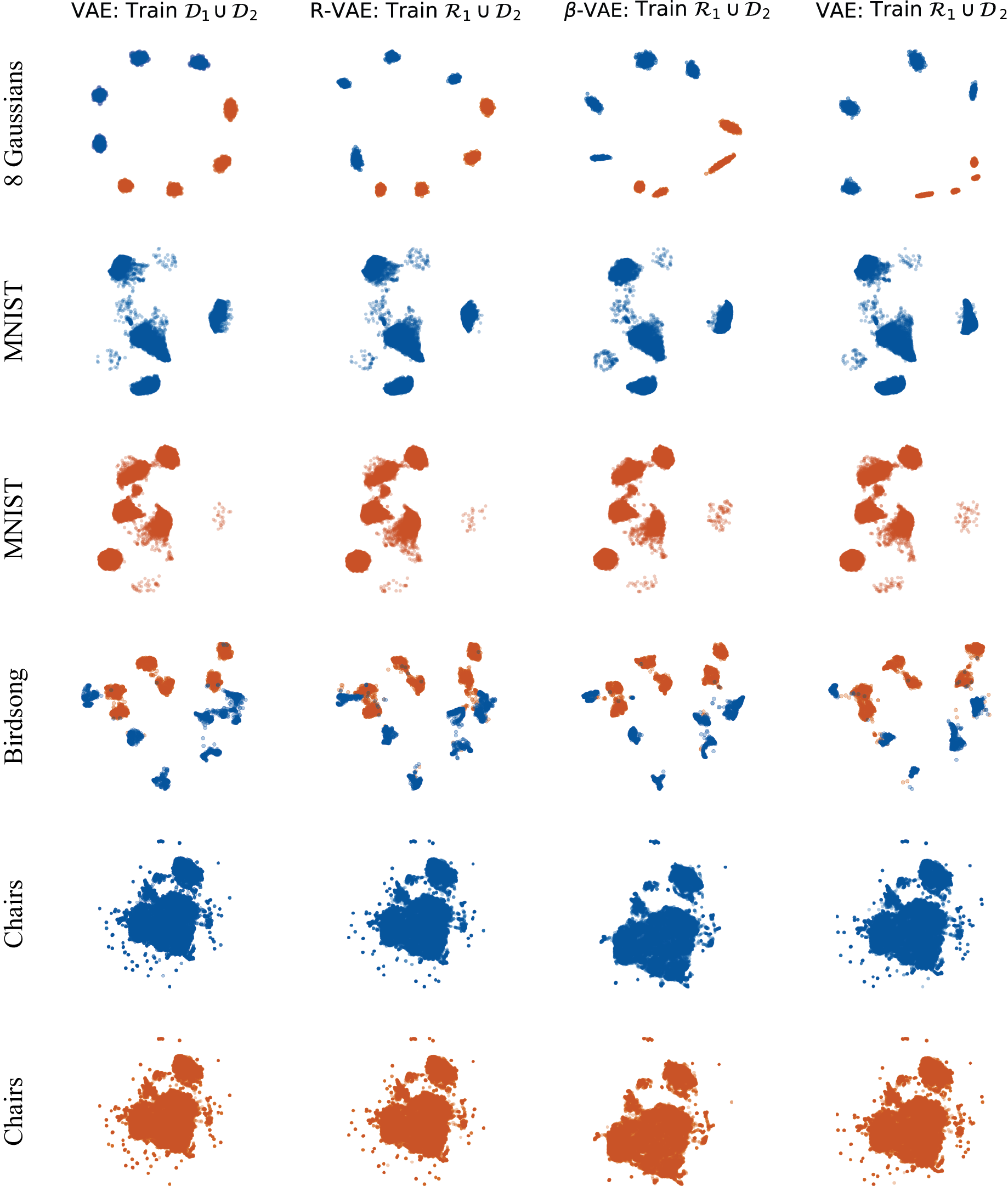}
    \caption{\textbf{Learned representations with linear correction.}  \textbf{First column:}Learned latent representations of each dataset, through joint training, projected to two dimensions by UMAP. \textbf{Second column:} Representations learned through training on $\mathcal{D}_1$ (used to find RPs). \textbf{Columns 2--4:} Latent spaces as reproduced by retraining with Rosetta VAE, standard VAE, and  $\beta$-VAE . $R=8,64,32, 128$ Rosetta points used for 8 Gaussians, MNIST, birdsong, and 3D Chairs respectively. In each case, the recovered latent space looks similar to the original target, suggesting rough linear equivalence between all learned models.}
    \label{fig:sf0}
\end{figure}

\section{Sequentially trained latent spaces uncorrected for linear transformation}
\label{app:sequential_latents}
\begin{figure}[H]
    \centering
    \includegraphics[width=\linewidth]{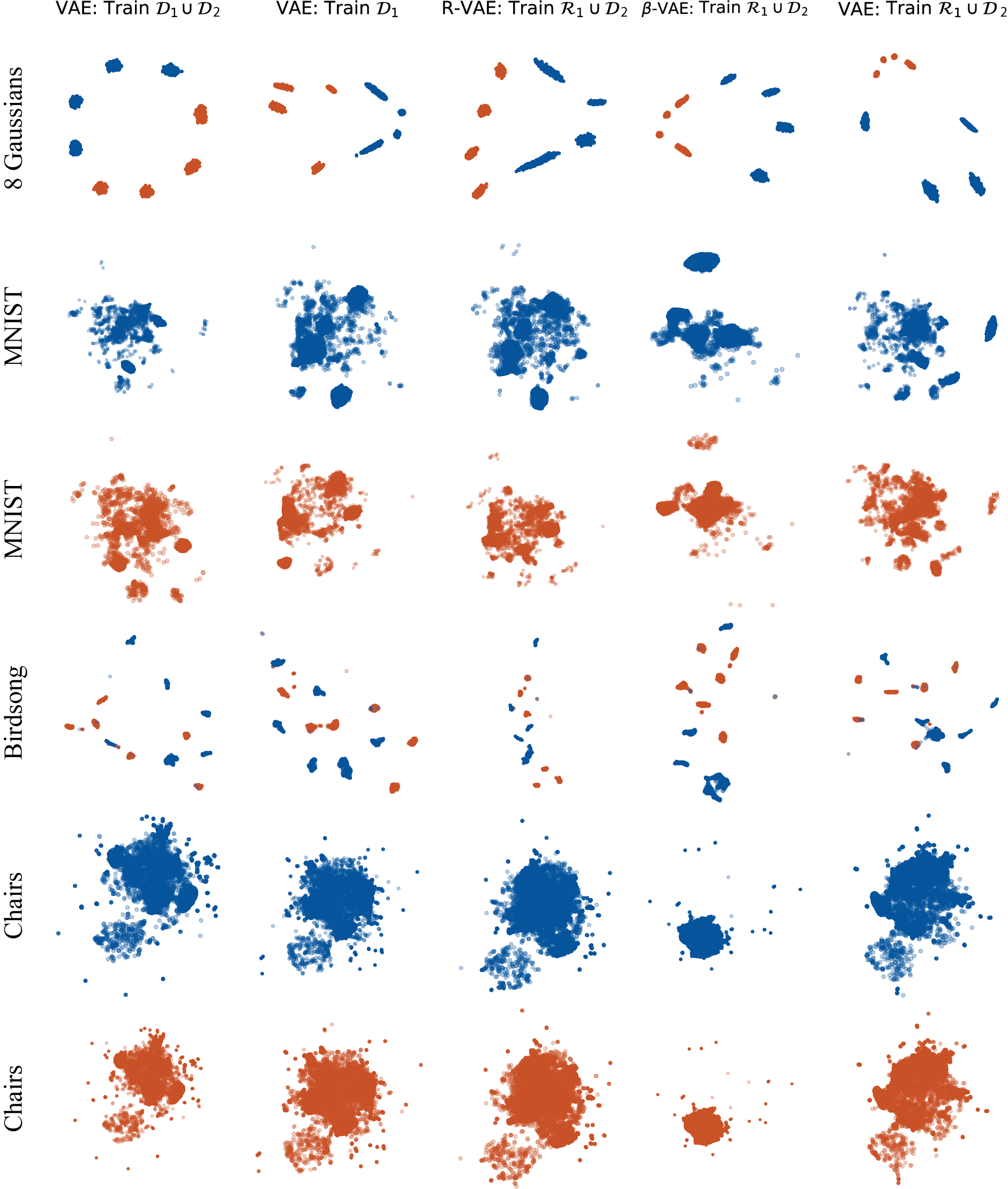}
    \caption{\textbf{Learned representations without linear correction.}  \textbf{First column:}Learned latent representations of each dataset, through joint training, projected to two dimensions by UMAP. \textbf{Second column:} Representations learned through training on $\mathcal{D}_1$ (used to find RPs). \textbf{Columns 2--4:} Latent spaces as reproduced by retraining with Rosetta VAE, standard VAE, and  $\beta$-VAE . $R=8,64,32, 128$ Rosetta points used for 8 Gaussians, MNIST, birdsong, and 3D Chairs respectively. For 8 Gaussians, the linear warping of the recovered latent spaces is obvious, while for MNIST and 3D Chairs, the R-VAE produces a less warpred version of the joint latent space than other models. For birdsong, the $\beta$-VAE appears least warped, as suggested by Table \ref{table:t2} and Figure \ref{fig:sf3}.}
    \label{fig:sf1}
\end{figure}

\section{Analysis of linear map in sequential case}
\label{app:mapAnalysis}
\begin{figure}[H]
    \centering
    \includegraphics[width=\linewidth]{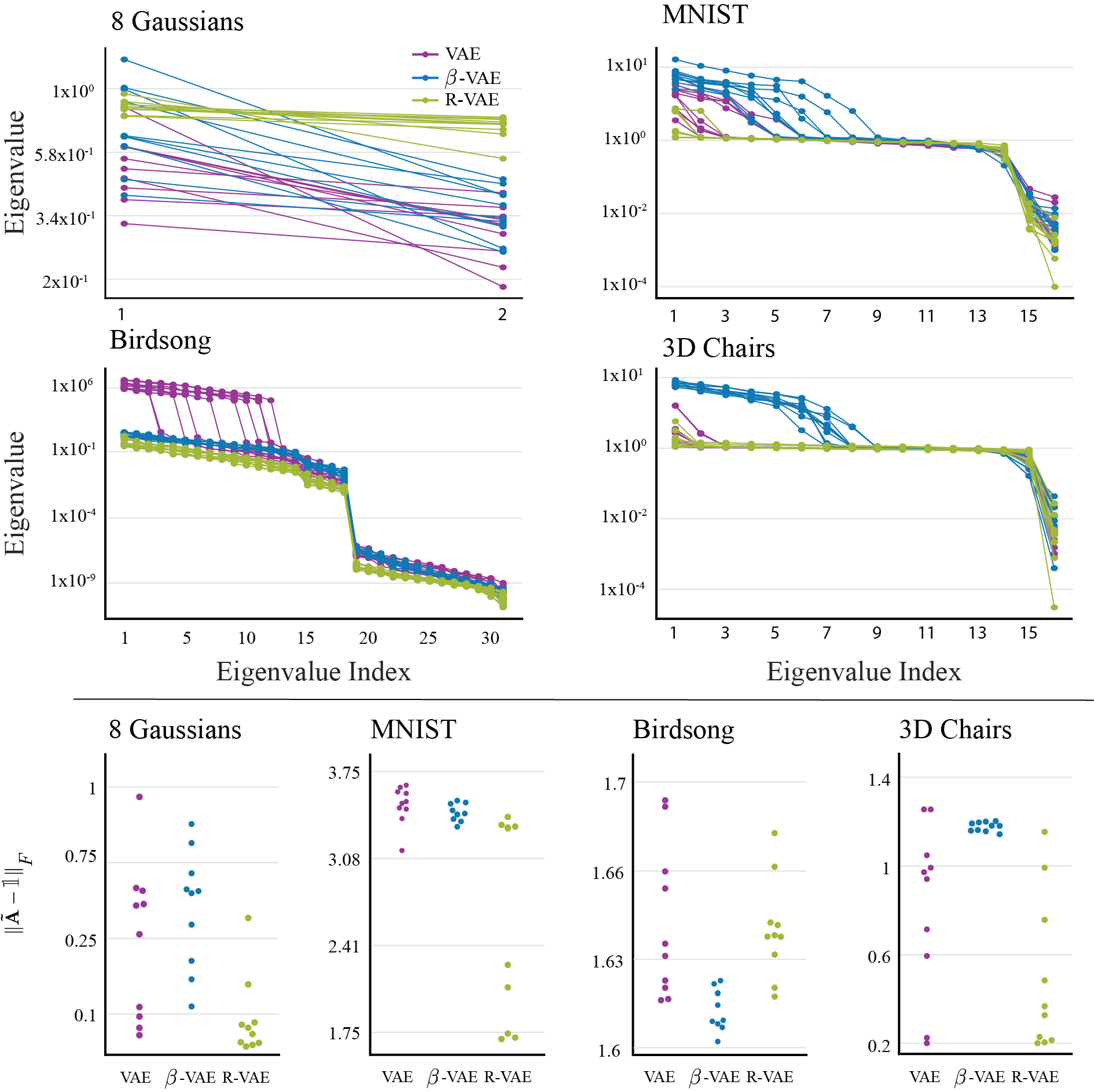}
    \caption{\textbf{Linear transformations between latent spaces are simplest for the R-VAE.} \textbf{First two rows:} Plots of the eigenvalues of the positive semidefinite matrix from polar decomposition of $\mathbf{A}$. In each case, the R-VAE exhibits the flattest spectrum, suggesting only a global rescaling without skew. \textbf{Bottom row:} Norm of the difference between the identity matrix and a version of $\mathbf{A}$ rescaled to have maximum singular value 1. The R-VAE has smaller values, indicating that less linear transformation is needed to align latent spaces found by sequential training.
   }
    \label{fig:sf3}
\end{figure}

To examine the latent embeddings further, we asked how distorted learned maps were across the various models and training runs by asking how close the learned $\mathbf{A}$ matrix in (\ref{eqn:distortion}) is to the identity. Models that best recapitulate the latent space should have $\mathbf{A} \approx \mathbbm{1}$. To investigate this, we performed a polar decomposition, $\mathbf{A} = \mathbf{U} \mathbf{P}$, with $\mathbf{U}$ an orthogonal matrix and $\mathbf{P}$ symmetric positive semi-definite. In Figure \ref{fig:sf3}, we plot the eigenvalue spectra for $\mathbf{P}$ for each model type. In all but the 8 Gaussians case, the spectra exhibit an abrupt drop, identifying an effective dimensionality for the data set. Moreover, in each case, the R-VAE exhibits the flattest spectrum, indicating less non-uniform stretching and compression of the learned space relative to the joint training template. We further investigate this by defining $\mathbf{\tilde{A}} = \mathbf{A}/\lVert \mathbf{A}\rVert_\infty$ to be the linear transformation with maximum rescaling of 1. That is, if the two latent spaces are the same up to a global rescaling, we should expect $\mathbf{\tilde{A}} \approx \mathbbm{1}$. In the bottom row of Figure \ref{fig:sf3}, we show the quality of this approximation across different models and training runs. Just as in Table \ref{table:t2}, the R-VAE shows lower distortion than both the VAE and $\beta$-VAE with the exception of birdsong, where the results are comparable.

\section{Learned biases of linear transformation are small}
\label{app:learned_biases}
\begin{figure}[H]
    \centering
    \includegraphics[width=\linewidth]{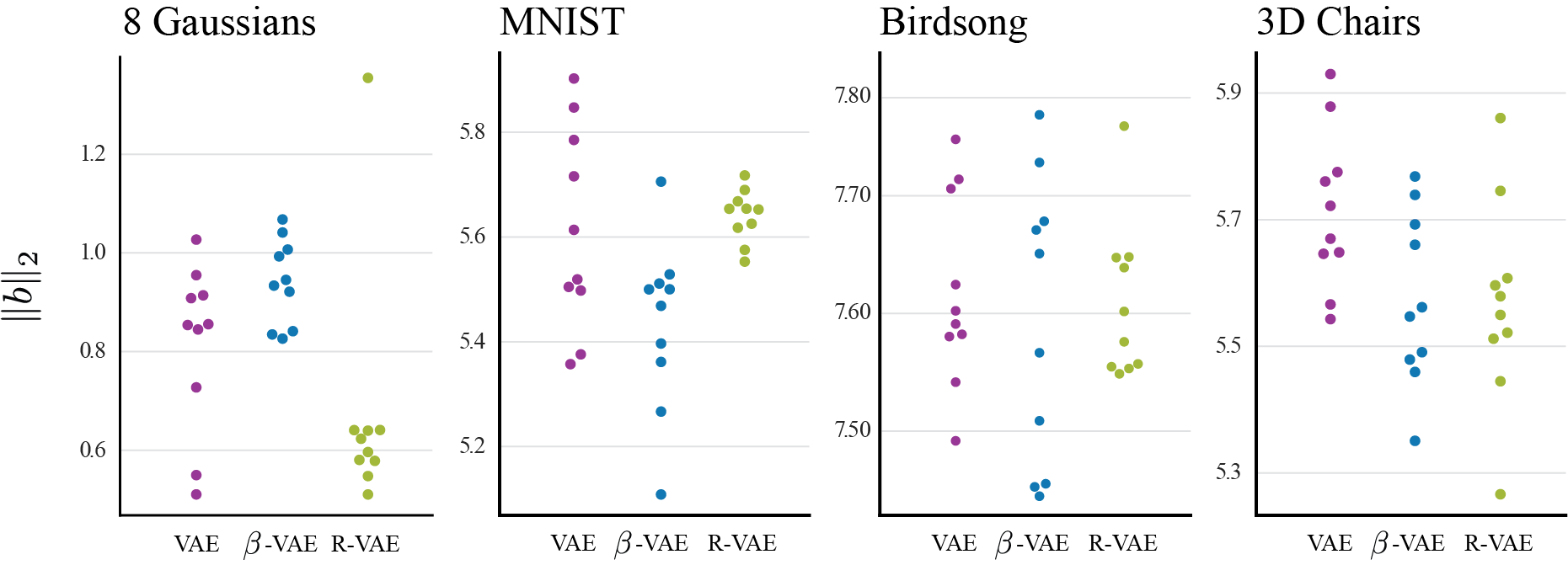}
    \caption{\textbf{Biases of linear transformations between latent spaces are small.} Norm of the learned bias vector for the linear transformation in (\ref{eqn:distortion}). The bias learned by for R-VAE tends to be smaller than learned by VAE and $\beta$-VAE, indicating that less linear transformation is needed to align latent spaces found by sequential training, although all biases are small.
   }
    \label{fig:sf2}
\end{figure}

\end{document}